\documentclass{article}

    \PassOptionsToPackage{round}{natbib}
    \PassOptionsToPackage{dvipsnames,svgnames,x11names}{xcolor}    
    \usepackage[final]{neurips_2025}

\pdfoutput=1  

\usepackage[english]{babel}
\usepackage{hyperref}
\usepackage{url}
\usepackage[utf8]{inputenc} 
\usepackage[T1]{fontenc}    
\usepackage{booktabs}       
\usepackage{amsfonts}       
\usepackage{nicefrac}       
\usepackage{microtype}      
\usepackage{amsmath}
\usepackage{amssymb}
\usepackage{placeins}
\usepackage{amsthm}
\usepackage{array}
\usepackage{enumitem}
\usepackage{subcaption}
\usepackage{dsfont}
\usepackage[export]{adjustbox}
\usepackage{multirow}
\usepackage{mycommands}

\hypersetup{
  colorlinks,
  citecolor=RoyalBlue,
  linkcolor=ForestGreen,
  urlcolor={blue!80!black},
}

\usepackage{cleveref}
\usepackage{algorithm, algpseudocode}
\usepackage{xcolor}

\usepackage{pifont}
\newcommand{\cmark}{\ding{51}}%
\newcommand{\xmark}{\ding{55}}

\long\def\comment#1{}

\def\energy{\ensuremath U_\theta}
\def\score{\ensuremath s_\theta}
\def\dim{\ensuremath d_{\rm eff}}

\title{Learning normalized image densities\\[0.2ex] via dual score matching}

\author{%
Florentin Guth \\
Center for Data Science, New York University \\
Flatiron Institute, Simons Foundation \\
\texttt{florentin.guth@nyu.edu}
\And
Zahra Kadkhodaie \\
Flatiron Institute, Simons Foundation \\
\texttt{zk388@nyu.edu}
\And
Eero P.~Simoncelli \\
New York University \\
Flatiron Institute, Simons Foundation \\
\texttt{eero.simoncelli@nyu.edu}
}

\begin{document}

\maketitle

\begin{abstract}
Learning probability models from data is at the heart of many machine learning endeavors, but is notoriously difficult due to the curse of dimensionality. We introduce a new framework for learning \emph{normalized} energy (log probability) models that is inspired by diffusion generative models, which rely on networks optimized to estimate the score. We modify a score network architecture to compute an energy while preserving its inductive biases. The gradient of this energy network with respect to its input image is the score of the learned density, which can be optimized using a denoising objective. Importantly, the gradient with respect to the noise level provides an additional score that can be optimized with a novel secondary objective, ensuring consistent and normalized energies across noise levels. We train an energy network with this \emph{dual} score matching objective on the ImageNet64 dataset, and obtain a cross-entropy (negative log likelihood) value comparable to the state of the art. We further validate our approach by showing that our energy model \emph{strongly generalizes}: log probabilities estimated with two networks trained on non-overlapping data subsets are nearly identical. Finally, we demonstrate that both image probability and dimensionality of local neighborhoods vary substantially depending on image content, in contrast with conventional assumptions such as concentration of measure or support on a low-dimensional manifold.
\end{abstract}

\section{Introduction}

Many problems in image processing and computer vision rely, explicitly or implicitly, on prior probability models. However, learning such models by maximizing the likelihood of a set of training images is difficult. The dimensionality of the space (i.e., the number of image pixels) is large, and worst-case data requirements for estimation grow exponentially (the ``curse of dimensionality'').  The machine learning community has developed a variety of methods to train a parametric network to estimate $\log p(x)$, known as an ``energy model'' \citep{hinton1986learning,lecun2006tutorial-ebm}, relying on the inductive biases of the network to alleviate the data requirements. For all but the simplest of models, this approach is frustrated by the intractability of estimating the normalization constant \citep{hinton2002contrastive-divergence,lecun2005loss-discriminative-ebm,yedidia2005bethe-approximation-bp,gutmann-hyvarinen2010noise-contrastive-estimation-nce,dinh2014nice,rezende2015variational,dinh2017density,song-kingma-ebm-tutorial}. 

A clever means of escaping this conundrum is to estimate the gradient of the energy with respect to the image (known as the ``score''), which eliminates the normalization constant, and can be learned from data with a ``score-matching'' objective \citep{hyvarinen2005estimation}.
The recent development of ``diffusion'' generative models \citep{sohlDickstein15,song2019generative,ho2020denoising,kadkhodaie2020solving} builds on this concept, by estimating a {\em family} of score functions for images corrupted by Gaussian white noise at different amplitudes. 
These scores may then be used to draw samples from the corresponding estimated density, using an iterative reverse diffusion procedure. These methods have enabled dramatic improvements in both the quality and diversity of generated image samples, but the learned density is implicit. An explicit and normalized energy model (and density) can be obtained through integration of the divergence of the score vector field along a trajectory \citep{song2020score}, at tractable but substantial computational cost.

Here, we leverage the power of diffusion models to develop a robust and efficient framework for directly learning a normalized energy model from image data.
We approximate the energy with a deep network that takes as input both a noisy image and the corresponding noise variance, and derive two separate objectives. The first, obtained by differentiating with respect to the noisy image, is a denoising objective (as used in diffusion models). The second, obtained by differentiating with respect to the noise variance, ensures consistency of the energy estimates across noise levels, which we show to be critical for obtaining accurate and normalized energies. 
We optimize the sum of the two, in a procedure that we refer to as ``\emph{dual} score matching''.
We also propose a novel architecture for energy, computed as the inner product of the input and output images of a score network. 
This preserves the inductive biases of the base score network, leading to equal or superior denoising performance (and thus, sample quality).

We train our energy model on ImageNet64 \citep{russakovsky2015imagenet,chrabaszcz2017downsampled-imagenet}, and show that the estimated energies 
lead to negative log likelihood (or cross-entropy) values comparable to the state of the art.
We further demonstrate that the energy model strongly generalizes in the sense of \citet{kadkhodaie2024generalization}: two separate models trained on \emph{non-overlapping} subsets of the training data assign essentially the \emph{same} probabilities to each image. This convergence is observed for training set sizes far smaller than the worst case prediction of the curse of dimensionality.
We find that the distribution of log probabilities over ImageNet images covers a broad range, with densely textured images at the lower end, and sparse images at the upper end. The probability is relatively insensitive to changes in image luminance, but decreases with dynamic range.  
Finally, we highlight two geometrical properties of the learned image distribution. The first is an extremely tight inverse relationship between volume and density that leads to an absence of concentration of energy values. They furthermore follow a Gumbel distribution, revealing an unexpected form of statistical regularity. The second is that the dimensionality of the energy landscape in the neighborhood of an image varies greatly depending on image content and neighborhood size. We observe images with full-dimensional neighborhoods of non-negligible size and images with lower-dimensional neighborhoods even at sub-quantization scales. These results challenge traditional presuppositions regarding high-dimensional distributions, such as the concentration of measure phenomenon \citep{vershynin2018high,wainwright2019high} and the manifold hypothesis \citep{tenenbaum2000global,bengio2013representation}.
Pre-trained models and software for running all experiments are available at \url{https://github.com/FlorentinGuth/DualScoreMatching}.

\comment{
\begin{itemize}
\item Classic density estimates: Gaussian, sparse wavelets, sparse overcomplete dictionaries, GSMs (?)
\item Early energy-based models: maximum likelihood, contrastive approaches (NCE), issues with normalization/sampling
\item Any push-forward or transport-based generative model (GANs, VAEs, NFs, diffusion models) leads to a density evaluated through change of variables formula. But very expensive, or limits expressivity (for NFs). Early normalizing flows to read: Rezende and Mohamed, Nice, realnvp (dinh et al), Glow
\item Auto-regressive approaches (Pixel-CNN and follow-ups)?
Chandler and field
\item Misc works on energy models that did not fit: Wand and Ponce, Sulam proximal prior
\end{itemize}
}

\section{Learning normalized energy models with dual score matching}
\label{sec:methods}

Estimating a high-dimensional probability density from samples faces two significant challenges as a result of the curse of dimensionality. The first is \emph{statistical}: realistically-sized datasets do not contain enough information to learn an unknown high-dimensional distribution. One thus needs powerful inductive biases, typically in the form of a parametric network architecture, to hope to recover the data distribution. The second challenge is \emph{computational}: the traditional objective aims to maximize the likelihood of the model over the data, which is intractable due to the need to compute the normalization constant. Nevertheless, recently developed generative models seem to succeed despite these problems. Drawing inspiration from diffusion models (\Cref{sec:motivation}), we develop a more direct solution by deriving a novel objective (\Cref{sec:objective}) and architecture (\Cref{sec:architecture}) for learning \emph{normalized} log probabilities (energies) from data. We validate our approach in \Cref{sec:performance}.

\subsection{Motivation}
\label{sec:motivation}

\begin{figure}
    \centering
    \includegraphics[height=4cm]{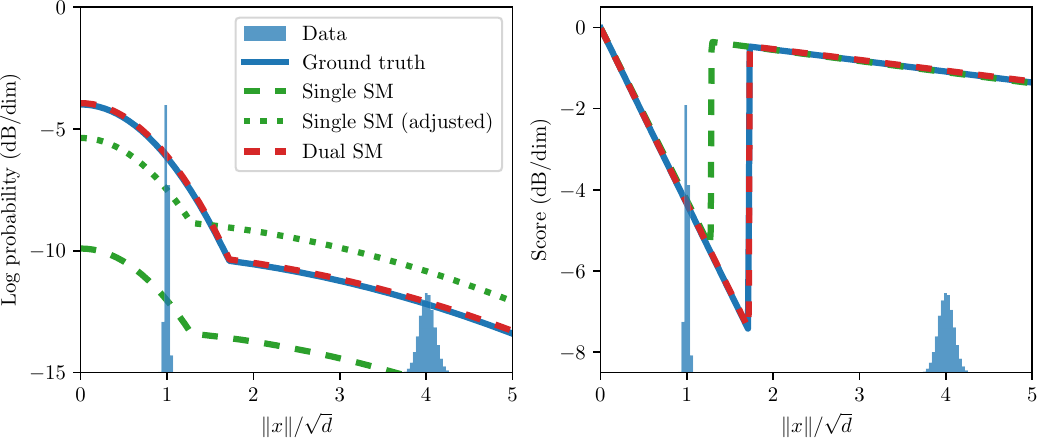}
    \hfill
    \includegraphics[height=4cm]{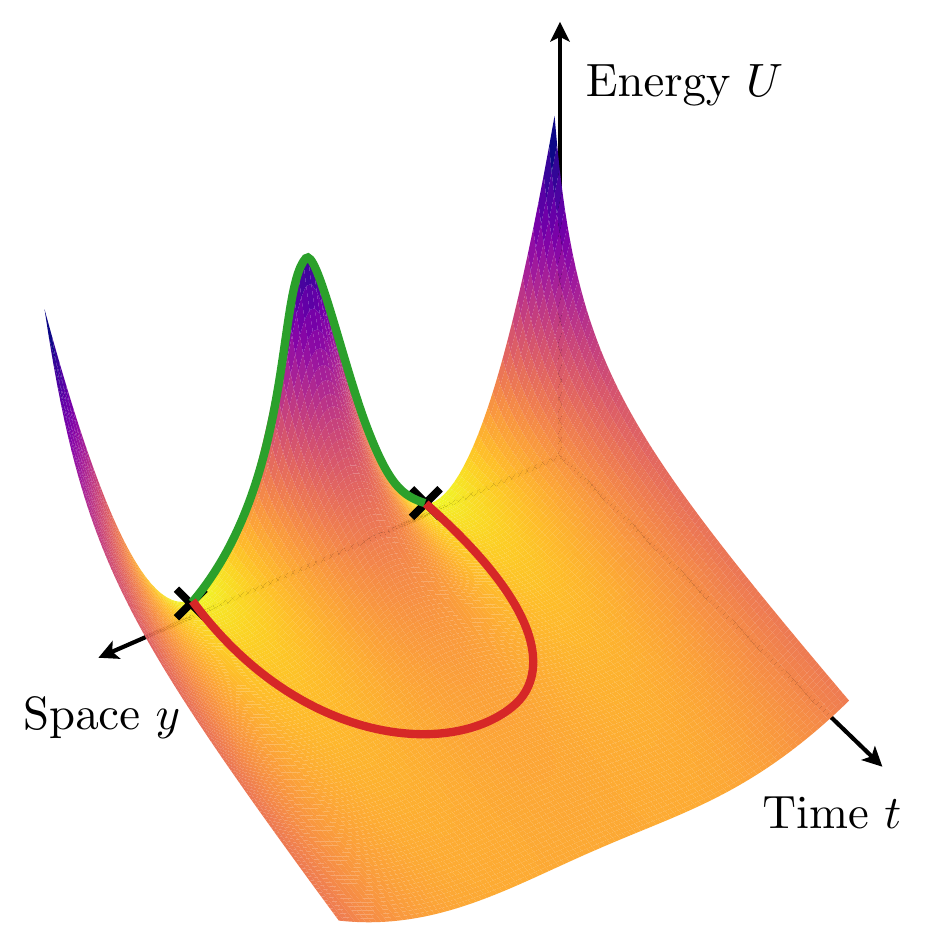}
    \caption{Comparison of single and dual score matching on recovering the energy of a scale mixture of two Gaussians in $d=1000$ dimensions. Experimental details are provided in \Cref{app:addtional_details}.
    \textbf{Left:} Radial slices of the log probability. The single score matching estimate (green dashed curve) fails to recover the true energy (blue solid curve), even after global normalization (green dotted curve), while dual score matching (red dashed curve) succeeds. 
    \textbf{Middle:} Radial components of the scores. Single score matching learns an accurate score over the support of the data (blue bar plot) but not outside of it.
    \textbf{Right:} Energy landscape across space and time (noise level) for a mixture of two Gaussians in one dimension. The direct path between the modes at $t=0$ crosses a large energy barrier (green curve), which is alleviated on a path that is not restricted to $t=0$ (red curve). 
    }
    \label{fig:single_vs_dual_sm}
    \vspace{-0.5em}
\end{figure}

Traditionally, energy models are defined in terms of a parametric function $\energy(x)$ that approximates the unnormalized log density over $x \in \RR^d$:  
$p_\theta(x) = \frac1{Z_\theta}\expe^{-\energy(x)}$, with a normalizing constant $Z_\theta = \int \expe^{-\energy(x)}\diff x$. The parameters $\theta$ are estimated by minimizing the expected negative log likelihood (NLL), $\expect[x]{-\log p_\theta(x)} = \expect[x]{\energy(x)} +\log Z_\theta$, which is equivalent to minimizing the KL divergence between $p_\theta(x)$ and the data distribution. 
The normalizing constant $Z_\theta$ plays a critical role in learning, representing the total energy over $\RR^d$ which trades off against the energy of the data. Unfortunately, direct estimation (i.e., computing the integral) is typically intractable.  

The normalization constant can be eliminated from the NLL by differentiating w.r.t.\ $x$, yielding a quantity known as the (negative) score: $-\nabla_x \log p_\theta(x) = \nabla_x \energy(x)$. As a result, a score model can be efficiently fitted to data via ``score matching'' \citep{hyvarinen2005estimation}, which minimizes the Fisher (as opposed to KL) divergence between the model and the data. For data corrupted by Gaussian white noise, this amounts to solving a denoising problem \citep{vincent2011connection,Raphan10}.
But this computational advantage comes at a statistical cost: a good approximation of the score does not always lead to a good model of the energy \citep{koehler2023statistical}, as we now illustrate. Consider the case of an equal mixture of two multivariate Gaussian distributions with zero mean and variances $\sigma_1^2$ and $\sigma_2^2$. In high dimensions, this distribution concentrates near the two hyper-spheres with radii $\sigma_1\sqrt d$ and $\sigma_2\sqrt d$, leading to data scarcity in the rest of the space.
We show in \Cref{fig:single_vs_dual_sm} energies estimated from samples by optimizing their gradient via (single) score matching. This fails to recover the true energy, even after adjusting a normalization constant. Indeed, evaluating the energy difference between the two Gaussians requires good estimation of the score \emph{along an integration path} between them. If there is no data in-between modes to constrain the score, due to an energy barrier or concentration phenomena, this leads to inconsistent energy values across modes. In other words, single score matching estimates energy values up to a \emph{mode-dependent} additive constant.

Diffusion models expand on score matching by learning scores of data corrupted by white Gaussian noise for a range of different noise amplitudes, $-\nabla_y U(y,t)$, with $y = x + \normal(0,t\Id)$. The evolution of the density $p(y|t)$ as ``time'' (noise variance) $t$ increases is a diffusion process, and thus the corresponding energies increase in smoothness with $t$. This \emph{multiscale} family of scores can be used to draw high-quality samples from $p(x)$ using a reverse diffusion algorithm, which follows a trajectory of partial denoising steps \citep{song2019generative,song2020score,ho2020denoising,kadkhodaie2020solving}. In fact, the diffusion scores implicitly capture a density model of the data \citep{song2021maximum,kingma2021variational-vdm}: the relative energy levels between modes, to which score matching at $t=0$ is blind \citep{zhang2022towards}, are encoded in the score at the time $t$ when they merge \citep{raya-ambrogioni2023spontaneous,biroli2024dynamical}. This implicit density model can be evaluated in various ways, which all involve a tractable but costly integration of score divergences (or denoising errors) \emph{as a function of time} \citep{song2020score,kong-ver-steeg-info-th-diffusion,skreta2024ito-density-estimation,karczewski2024diffusion-cartoonists}. See \Cref{app:density_evaluation_dm} for a more in-depth exposition. 

Intuitively, diffusion models deal with multimodal distributions by providing a high-probability path between modes through space \emph{and time}, as visualized in the right panel of \Cref{fig:single_vs_dual_sm}. In other words, the \emph{joint} distribution $p(y,t)$ qualitatively has a connected support. This suggests that we may learn the joint ``space-time'' energy $U(y,t)$ by score matching on $(y,t)$. Specifically, the ``space score'' $-\nabla_y \energy(y,t)$ can be learned using a denoising objective, and the ``time score'' $-\partial_t \energy(y,t)$ \citep{choi2022density} can be learned using an analogous score matching objective. Matching \emph{both} space and time scores correctly constrains the energy levels across modes: the energy difference between two modes can be explicitly recovered by integrating the energy derivative along the red path in \Cref{fig:single_vs_dual_sm} (although we will not need to do so explicitly).

\subsection{Dual score matching: Objective function}
\label{sec:objective}

\paragraph{Space and time score matching.}
We aim to learn a time-dependent energy function, $\energy(y,t)$, to approximate the NLL of the noisy image distribution at all noise levels $t$:
\begin{equation}
    U(y,t) = -\log\paren{\int p(x) \expe^{-\frac1{2t}\|x-y\|^2 - \frac{d}2\log\paren{2\pi t}} \diff x}.
\end{equation}
Differentiating this NLL with respect to $y$ gives the (negative) ``space score'', which can be expressed using the Miyasawa-Tweedie identity \citep{Robbins1956Empirical,Miyasawa61} as
\begin{equation}
    \nabla_y U(y,t) = \expect[x]{\frac{y-x}{t} \st y}.
\end{equation}
This leads to the denoising score matching objective \citep{vincent2011connection,Raphan10,saremi2018deep} used to train diffusion models:
\begin{equation} 
    \label{eq:dsm_loss}
    \ell_{\rm DSM}(\theta,t) = \expect[x,y]{\norm{\nabla_y \energy(y,t) - \frac{y-x}t}^2}.
\end{equation}
Differentiating the energy with respect to $t$ yields a (negative) ``time score'' \citep{choi2022density}, for which we can derive a similar identity (see \Cref{app:time_score_matching}):
\begin{equation}
    \label{eq:time_score_identity}
    \partial_t U(y,t) = \expect[x]{\frac{d}{2t} - \frac{\norm{y-x}^2}{2t^2} \st y}.
\end{equation}
This leads to an analogous ``time score matching'' objective:
\begin{equation}
    \label{eq:tsm_loss}
    \ell_{\rm TSM}(\theta,t) = \expect[x,y]{\paren{\partial_t \energy(y,t) - \frac{d}{2t} + \frac{\norm{y-x}^2}{2t^2}}^2} .
\end{equation}

\paragraph{Combining objectives across noise levels.}
The two objectives in \cref{eq:dsm_loss,eq:tsm_loss} are defined for a given noise level $t$.  We form an overall objective by integrating these over $t$, with an appropriate weighting. For the denoising score matching objective, a natural choice is the so-called \emph{maximum-likelihood weighting} \citep{song2021maximum,kingma2021variational-vdm}, which provides a bound on the KL divergence with the data distribution:
\begin{equation}
    \label{eq:ml_weighting}
    \kl{p}{\tilde p_\theta} \leq \frac12 \int_0^\infty \ell_{\rm DSM}(\theta,t) \, \diff t,
\end{equation}
where $\tilde p_\theta$ is the implicit density of the generative diffusion model.
This integral can be approximated by Monte-Carlo sampling from a distribution of noise levels. In practice, we have found it best to sample $\log t$ uniformly over a finite interval (specifically, $p(t) \propto 1/t, t \in [t_{\min}, t_{\max}]$), which corresponds to the following implementation of the integral:
\begin{equation}
    \int_{t_{\min}}^{t_{\max}} \ell_{\rm DSM}(\theta,t) \, \diff t 
    = \expect[t]{t \, \ell_{\rm DSM}(\theta,t)}.
\end{equation}
Note that the resulting term in the expected value is unit-less: it is invariant to simultaneous rescaling of the data and noise level. We thus choose to weight the time score matching objective in \cref{eq:tsm_loss} by $t^2$, so that it is also unit-less, and to evaluate it over the same distribution $p(t)$. Finally, after appropriate normalization by the dimensionality $d$ that ensures that the two objectives have comparable orders of magnitude, we simply add them (we found no significant improvement from tuning a tradeoff hyperparameter). In summary, our {\em dual score matching objective} is:
\begin{align}
    \label{eq:loss}
    \ell(\theta) &= \expect[t]{\frac td \, \ell_{\rm DSM}(\theta,t) + \paren{\frac td}^2 \, \ell_{\rm TSM}(\theta,t)}. 
\end{align}

\paragraph{Normalization.} Since we match only the partial derivatives of $U_\theta(y,t)$ to those of the true energy $U(y,t)$, we only recover the energy up to a \emph{global} constant: at the end of training, $U_\theta(y,t) \approx U(y,t) + \mathrm{const}$. Note again that this crucially relies on $p(y,t)$ having a connected support. An important aspect of our framework is that it enables estimation of this constant, which determines the normalization of $\expe^{-U_\theta(y,t)}$. Indeed, the time score objective ensures that this normalizing constant does not depend on time: \emph{mass is conserved through the diffusion}. Since the true distribution is approximately Gaussian at large noise levels, $p(y|t_{\max}) \approx \normal(0, t_{\max}\Id)$, we can set this constant to the entropy of this Gaussian distribution:
\begin{equation}
    \label{eq:normalization}
    \energy(y,t) \longrightarrow \energy(y,t) - \expect[y]{\energy(y,t)\st t=t_{\max}} + \frac{d}{2}\log(2\pi\expe t_{\max}).
\end{equation} 
\Cref{fig:single_vs_dual_sm} provides a high-dimensional numerical example verifying that dual score matching indeed provides an accurate estimate of \emph{normalized} energy values (in particular, for $t=0$), in contrast to single score matching. 

\paragraph{Related approaches.} Similar combinations of space and time scores were considered by \citet{choi2022density} (who referred to it as the ``pathwise'' method) and \citet{kobler2023learning}.
These time score objectives, however, relied on a second derivative in time instead of a regression objective. 
Our time score objective is also a special case of the conditional time score matching objective developed concurrently in \citet{yu2025density}. 
Finally, \citet{yadin2024classification} train an energy model with an objective combining score matching with a classification cross-entropy loss for estimating a discretized version of the noise level $t$.

\subsection{Dual score matching: Architecture}
\label{sec:architecture}

How should one choose an architecture to compute the energy $\energy(y,t)$? Rather than designing one from scratch \citep{salimans2021should,cohen2021has,thiry2024classification}, we construct one by modifying a score-based denoising architecture that is known to have appropriate inductive biases.
Let $\score \colon \RR^d \times \RR \to \RR^d$ be such an architecture (e.g., a UNet). We wish to define a new energy architecture $\energy \colon\RR^d \times \RR \to \RR$ such that $\nabla_y \energy \approx \score$, preserving the inductive biases of $\score$. 
To achieve this, we set
\begin{equation}
    \label{eq:energy_architecture}
    \energy(y,t) = \frac12 \inner{y, \score(y,t)}.
\end{equation}
This inner product parameterization has been concurrently proposed by \citet{thornton2025composition}.  We show in \Cref{app:energy_architecture} that $\nabla_y \energy(y,t) = \score(y,t)$ if the score network $\score$ is both conservative and \emph{homogeneous} \citep{romano-elad-milanfar-red,reehorst2018regularization}. Homogeneity has been shown to hold approximately in the related setting of blind denoisers \citep{MohanKadkhodaie19b,herbreteau2023normalization}, and can be enforced architecturally. Note that the seemingly similar choice of a squared norm architecture $\energy(y,t) = \frac12\norm{\score(y,t)}^2$ used in some previous energy models \citep{salimans2021should,hurault2021gradient,du2023reduce-reuse-recycle,yadin2024classification,thiry2024classification} leads to the same desirable homogeneity properties in $y$ \emph{but not in $s_\theta$}, and thus fails to preserve the optimization properties of the original score network. 
Further architectural details are provided in \Cref{app:architecture_details}.


\subsection{Performance evaluation}
\label{sec:performance}

\paragraph{Denoising performance.}
We first verify that the gradient of the proposed energy network, $\nabla_y \energy$, provides as good a denoiser as the score network $\score$ on which it is based. 
We train the two networks separately, $\energy$ with the dual score matching objective (\cref{eq:loss} using double back-propagation), and $s_\theta$ with the standard (denoising) score matching objective (\cref{eq:dsm_loss}). Further details are provided in \Cref{app:training_details}. \Cref{tab:denoising_performance} compares denoising performance across noise levels.
For all but the smallest noise levels, the energy-based model achieves (slightly) better denoising performance than the score model. Thus, there is no penalty in modeling the energy rather than the score. This is contrary to the results of \citet{salimans2021should}, and is likely due to our use of an architecture that is homogeneous, and for which non-conservativeness of the score is not a large source of denoising error \citep{MohanKadkhodaie19b,chao2023investigating}. Finally, these results demonstrate that the two components of our dual score matching objective are not trading off against each other; rather, they complement and even \emph{reinforce} each other (i.e. their minima coincide).


\begin{table}
    \centering
    \caption{Denoising performance at several noise levels for corresponding score and energy networks, averaged over images in ImageNet64, with intensities scaled to lie in $[0,1]$. All quantities are expressed as peak signal-to-noise ratio (PSNR) in dB: $\mathrm{PSNR} = -10\log_{10}(\mathrm{MSE})$.}
    \vspace{0.5em}
    \label{tab:denoising_performance}
    \small
    \begin{tabular}{rccccccccc}
        \toprule
        Noise variance & $90$ & $75$ & $60$ & $45$ & $30$ & $15$ & $0$ & $-15$ & $-30$ \\
        \midrule
        Score network & $\mathbf{90.20}$ & $\mathbf{75.47}$ & $60.43$ & $47.19$ & $35.58$ & $25.92$ & $18.84$ & $13.31$ & $-0.11$ \\
            Energy network & $90.09$ & $75.17$ & $\mathbf{60.45}$ & $\mathbf{47.25}$ & $\mathbf{35.67}$ & $\mathbf{26.01}$ & $\mathbf{18.88}$ & $\mathbf{13.48}$ & $\mathbf{\hphantom{-}2.53}$ \\
        \bottomrule
    \end{tabular}
\end{table}

\paragraph{Negative log likelihood.}
How accurate are our energy estimates? A standard evaluation of probabilistic models consists of estimating the cross-entropy $\expect[x]{-\log p_\theta(x)}$ between the data distribution and the model, also known as negative log likelihood (NLL). In practice, NLL is computed by computing the probability the model assigns to a held-out test set. There are however several subtleties that make direct model comparisons challenging. Specifically, there are 3 factors that can cause variations in NLL of $\pm 0.5$ bits/dimension or more: details of data pre-processing (e.g., downsampling or data augmentation), conversion method from continuous to discrete probability, and the estimator type (exactly normalized, variational bound, or approximately normalized). We expand on these issues in \Cref{app:nll}.

We compare NLLs of our method and a variety of recent energy models in \Cref{tab:nll}. This evaluation demonstrates that our model is comparable to the best-performing models in the literature, within the variability arising from the three factors of variation mentioned in the previous paragraph. 
Two unique advantages of our method are that it provides (1) direct (one-shot) estimates of energy (as opposed to other density estimation approaches used with diffusion models, which we review in \Cref{app:density_evaluation_dm}), and (2) access to energy across all noise levels, providing a window into larger-scale features of the energy landscape. For instance, our energy network can compute the probability of 50k images in 12s on an A100 GPU, whereas the score network requires upwards of 3h20, even with as few as 100 times steps and 10 noise samples to compute the energy (see \Cref{app:density_evaluation_dm}). This justifies the longer training time due to the cost of double-backpropagation (training for 1M steps on ImageNet64 on a single A100 GPU respectively took 120 hours, compared to 32 hours for the score network), which may be alleviated by the use of sliced score matching \citep{song2020sliced}.


\begin{table}
    \vspace{-0.5em}
    \centering
    \caption{Negative log likelihood (in bits/dimension) on ImageNet64 test set. See \Cref{app:nll} for more details.}
    \vspace{0.5em}
    \label{tab:nll}
    \scriptsize
    \begin{tabular}{rcccccr}
        \toprule
        Method & Anti-aliasing & Augmentation & Discreteness & Type & Single NFE & NLL \\
        \midrule
        Glow \citep{kingma2018glow} & \xmark & None & Continuous & Normalized & \cmark & 3.81 \\
        PixelCNN \citep{van-den-oord2016pixel-cnn} & \xmark & None & Discrete & Normalized & \xmark & 3.57 \\
        I-DDPM \citep{nichol-dhariwal2021improved-ddpm} & \xmark & None & Continuous & Upper bound & \xmark & 3.54 \\
        VDM \citep{kingma2021variational-vdm} & \xmark & None & Discrete & Upper bound & \xmark & 3.40 \\
        FM \citep{lipman2023flow-matching} & \cmark & None & Uniform* & Normalized & \xmark & 3.31 \\
        NFDM \citep{bartosh2024neural-nfdm} & \xmark & Horizontal flips & Uniform* & Normalized & \xmark & 3.20 \\
        TarFlow \citep{zhai2024tarflow} & \xmark & Horizontal flips & Uniform & Normalized & \cmark & 2.99 \\
        \midrule
        Ours & \cmark & Horizontal flips & Continuous & Estimate & \cmark & 3.36 \\
        \bottomrule
    \end{tabular}
\end{table}

\section{Analysis of the learned energy-based model}

\subsection{Generalization}

\begin{figure}
    \centering
    \includegraphics[width=\linewidth]{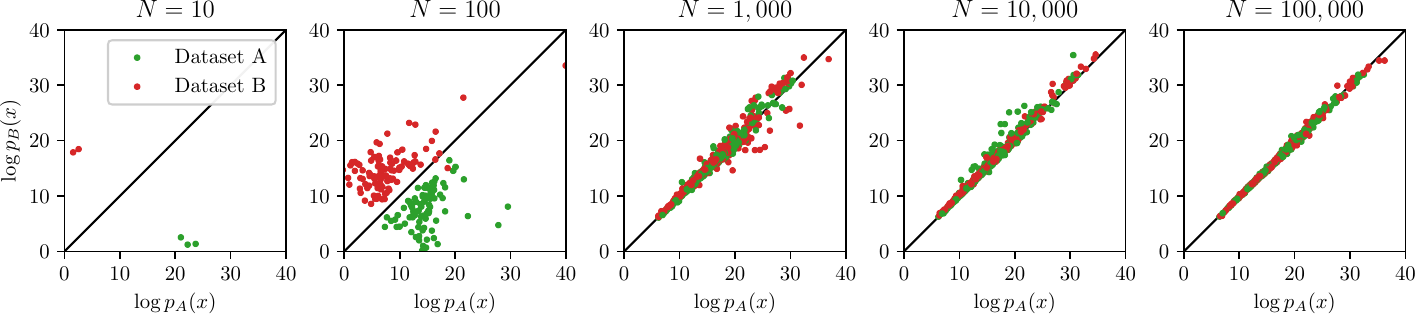}
    \caption{Convergence of energy estimates. The data set is split into two halves (denoted A and B), and separate energy models are trained on $N$ samples drawn from each half. Each scatterplot compares the energy estimates of the two models at $t=0$, over all $2N$ training images. As $N$ increases, the energy estimates of the two models converges for all images. 
    }
    \label{fig:generalization}
\end{figure}

The previous section demonstrated that our energy-based model achieves near state-of-the-art NLL on ImageNet64. That is, the model \emph{on average} assigns high probability to a set of held-out test images. Next, we establish that the energies of the \emph{individual} images are reliable. In particular, we verify that the model's energy assignment is stable under change of the training data. 

To this end, we use the strong generalization test developed in \citet{kadkhodaie2024generalization}. We partition the training data into two non-overlapping sets, train a separate energy-based model on each set, and then compare the energies computed by these two models on images from both training subsets. We gradually increase the size of each training set until the two models assign approximately equal log probability across all images. \Cref{fig:generalization} shows the results of this experiment. The two models assign very different probabilities to the same image when the training set size, $N$, is small. But they converge gradually and compute nearly the same values at $N=10^5$. Note that the rate of convergence depends on image probability: more data is needed before the two models agree on the high-probability images. It is also worth noting that the transition from memorization to generalization of energy models is marked by a large increase of variance in energy over the training set (starting at $N=100$).

This result establishes that the model variance vanishes with a feasible training set size. However, it does not guarantee that the values are accurate---the models could be biased. 
Direct calculation of model bias requires access to the true density, which is only available for synthetic data (as in \Cref{fig:single_vs_dual_sm}). However, the small NLL values over test data (\Cref{tab:nll}) do suggest a small model bias.



\subsection{Distribution of energies and relationship to image content}
\label{subsec:distribution}
\begin{figure}
    \centering
    \includegraphics[width=0.47\linewidth]{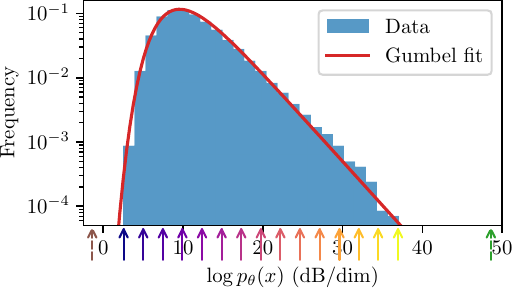}
    \hfill
    \raisebox{4pt}{\includegraphics[width=0.43\linewidth]{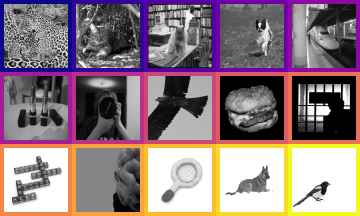}}
    \caption{
    Histogram of log probabilities of test images in the ImageNet dataset. Color-coded arrows indicate values for the example images on the right, and the leftmost (brown) and rightmost (green) arrows indicate values for a uniform noise image in $[0,1]$ and a constant image of intensity $0.5$, respectively. The distribution is well-fit by a Gumbel distribution (red line). Additional example images organized by probability are provided in \Cref{fig:low-energy-images,fig:high-energy-images,fig:linear-energy-images} (\Cref{app:more_images}).
}
    \label{fig:energy_distribution}
\end{figure}

We now study properties of the learned energy model. What is the entropy (average energy) of the image dataset? Do all images have the same probability? If not, what determines the probability of an image? To investigate these questions, we compute the log probability of all $50,000$ images in the ImageNet64 test set. We express log probability in units of decibels per dimension (dB/dim), computed as $\frac1d 10\log_{10} p_\theta(x)$, with $d = 64 \times 64 = 4096$. In this scale, an increase of $10$ dB/dim corresponds to a multiplicative rescaling of the probability density by a factor of $10^d$.

\paragraph{Entropy.}
The average value of $-\log p_\theta(x)$ provides an estimate of $-11.4$ dB/dim for the differential entropy of ImageNet. Note that the uniform distribution over the hypercube $[0,1]^d$ has an entropy of $0$. This indicates that natural images occupy a fractional volume of about $10^{-1.14d}$. This can be converted to an estimate of discrete entropy by assuming that log probability is constant within quantization bins. For $8$-bit images (with $256$ possible intensity values), this corresponds to an entropy of $4.20$ bits/dimension (the deviation from the value in \Cref{tab:nll} is due to the use of grayscale images here). In other words, there are $\sim 10^{5,180}$ quantized ImageNet images out of $10^{9,860}$ possible images at this resolution.

\paragraph{Lack of concentration.}
Many high-dimensional probability distributions exhibit a \emph{concentration} phenomenon: typical realizations have nearly equal energy. For instance, in simple image probability models such as a Gaussian model or a sparse wavelet model (where wavelet coefficients are independent), the energy of independent components add, and the law of large numbers imply a concentration of the energy around its mean. 
In contrast, our energy network reveals enormous diversity in the log probabilities of individual ImageNet images (\Cref{fig:energy_distribution}, left). Over the entire test set, they span a range of $34.4$ dB/dim, corresponding to a probability ratio of $\approx 10^{14,000}$. Given this enormous ratio, it is surprising that any low probability image appears in a finite dataset. To make sense of this apparent contradiction, it is important to realize that the total probability mass of images with a given value of $\log p$ in a dataset is equal to this probability value \emph{multiplied by the volume of the corresponding probability level set}. Observing a range of probability values of $10^{14,000}$ thus reveals that these enormous variations in probability density must be nearly compensated for by inverse variations in the volume of their corresponding level sets.

\paragraph{Shape of the distribution of log probabilities.}
The distribution of log probability values is highly skewed (\Cref{fig:energy_distribution}, left): there is a heavy tail of high-probability images and a much lighter tail of low-probability images.
We also observed these qualitative features on the CelebA and CIFAR-10 datasets, shown in \Cref{app:more_histograms}. 
Surprisingly, this distribution is well approximated by a Gumbel distribution (we report parameter values in \Cref{app:addtional_details}), which arises as a limit distribution for the maximum of many i.i.d.\ random variables. This surprising observation is, to the best of our knowledge, novel, and calls for an explanation. 
In independent component models, the log probabilities are sums of i.i.d.\ random variables, and are therefore Gaussian-distributed in high dimensions.
In contrast, a simple model that leads to a Gumbel distribution is a high-dimensional spherically-symmetric distribution with an exponentially-distributed radial marginal \citep{Lyu08c}.

\begin{figure}
    \centering
    \includegraphics[width=0.32\linewidth]{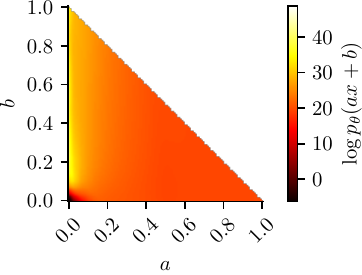}
    \hfill
    \includegraphics[width=0.30\linewidth]{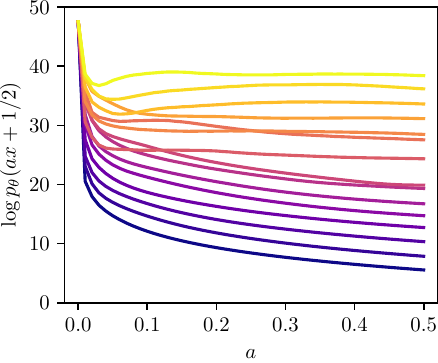}
    \hfill
    \includegraphics[width=0.30\linewidth]{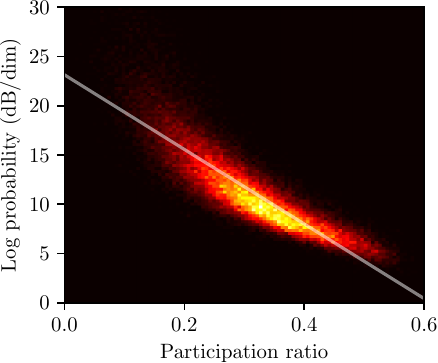}    
    \caption{Influence of image statistics on probability. \textbf{Left.} $\log p_\theta(a x + b\comment{\mathbf{1}})$ as a function of $a$ and $b$. \textbf{Middle.} Horizontal slice ($b = \frac12$) of the left panel for the example images of \Cref{fig:energy_distribution}.
    \textbf{Right.} Log probability as a function of sparsity, measured as the participation ratio of wavelet coefficients.}
    \label{fig:energy_vs_sparsity}
\end{figure}

\paragraph{Image content.}
We also show in \Cref{fig:energy_distribution} (right) example images with a variety of log probabilities. High-probability images contain small objects, on a blank (often white) background. Conversely, low-probability images are filled with dense detailed texture. Since the energy of an image, when expressed in bits, corresponds to the size of its optimally compressed representation, this implies that higher-probability images have shorter codes and thus less ``information content''. We show in \Cref{app:logp_optimization} images that have been obtained by locally maximizing or minimizing probability.

\paragraph{Intensity range and sparsity.}
Following these visual observations, we further examine the influence of intensity range and sparsity on image probability (\Cref{fig:energy_vs_sparsity}). First, we evaluate the log probability of reference images as we manipulate their brightness and contrast through an affine operation. We find that the brightness has minimal effect on the probability, while higher-contrast images have lower probability. This reveals that the distribution support is \emph{star-shaped}: any pair of images are connected by a high-probability path passing through a constant image. In contrast, we show in \Cref{app:interpolation} that linear interpolation between pairs of images typically increase probability: the distribution support is not convex.
Next, we verify that log probability is correlated with a simple measure of sparsity of images. We compute a multi-scale wavelet decomposition of the images, and measure the $\ell^1$-norm divided by the $\ell^2$-norm of the coefficients. The square of this quantity, $\Vert x\Vert_1^2/ d \Vert x \Vert^2_2 \in (0,1]$ is known as the participation ratio, with smaller values indicating higher sparsity. The right panel of \Cref{fig:energy_vs_sparsity} shows that this simple measure captures a significant portion of the variance of $\log p_\theta(x)$. 

\subsection{Effective dimensionality of the energy landscape}

Beyond estimating log probability for a given image, we aim to characterize the local behavior of the density in the vicinity of that image. For instance, we would like to assess whether the probability is locally concentrated near a low-dimensional ``tangent'' subspace, and if so, to estimate its dimensionality. In this section, we explain how these quantities may be computed from our energy model. We then show that the local dimensionality around an image is often much lower than the ambient dimensionality of the space, but that this dimensionality depends on both the particular image and the scale that is used to define the local neighborhood.

\begin{figure}
    \centering
    \raisebox{1.5ex}{\includegraphics[width=0.43\linewidth]{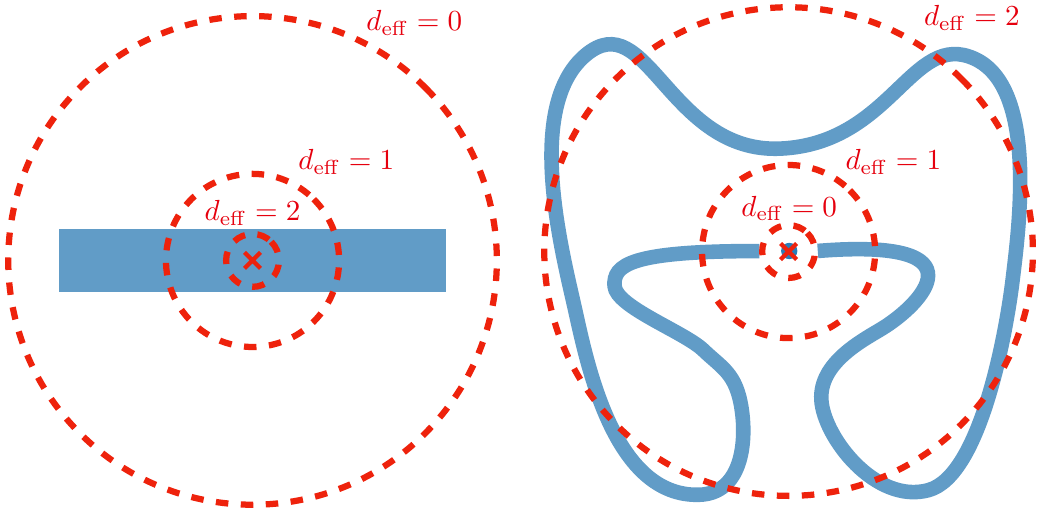}}
    \hfill
    \includegraphics[width=0.53\linewidth]{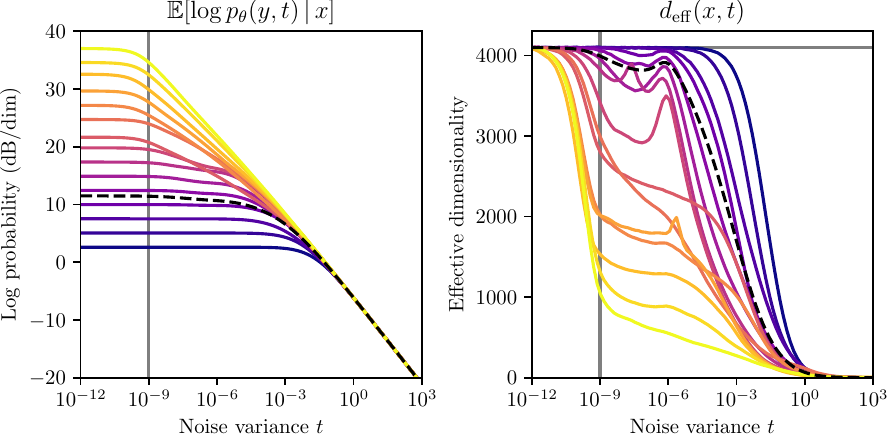}
    \caption{
    \textbf{Left:} Two hypothetical examples illustrating how local effective dimensionality depends on the scale of the neighborhood. For both examples, support of density corresponds to the blue regions. In the left example, the dimensionality around the red point decreases with scale (from 2 down to 0), while the opposite is true for the right example.
    \textbf{Right:}
    Log probability and effective dimensionality as a function of noise level. Colored lines correspond to different example images $x$ (shown in right panel of \Cref{fig:energy_distribution}), while the dashed black line shows the average over the ImageNet test set. The vertical gray line indicates the minimum noise level presented during training ($t = 10^{-9})$, and the horizontal gray line the ambient dimensionality of the dataset ($d = 4096$).}
    \label{fig:dimensionality}
\end{figure}

\paragraph{Multi-scale dimensionality.}
Consider hypothetical distributions supported on the blue regions in the left two panels of \Cref{fig:dimensionality}. For the left example, the effective dimensionality of a local neighborhood decreases with the size of that neighborhood. 
The opposite behavior is also possible, as shown in the right example. These examples demonstrate that dimensionality measures which aim to describe these geometrical structures need to depend on both the location and the scale of the neighborhood \citep{MohanKadkhodaie19b,tempczyk2022lidl-dimensionality}.


\paragraph{Effective dimensionality.}
We now introduce an effective dimensionality measure which can be equivalently defined from the optimal denoiser or the evolution of the probability landscape as it diffuses. Given a noisy observation $y$ of a clean image $x$ with noise variance $t$, the optimal denoiser estimates $x$ with the conditional expectation $\expect{x\st y}$. Its average deviation from $x$ capture the local support of the data distribution around $x$ at scale $t$. 
Intuitively, from observing $y$, the optimal denoiser identifies that $x$ is located on a $\dim$-dimensional ``tangent'' space and projects $y$ onto it. This preserves the components of the noise that lie along the subspace, incurring a denoising error $t\dim$.
Thus, we define the local effective dimensionality around $x$ at scale $t$ as
\begin{equation}
    \label{eq:dim_mse}
    \dim(x,t) = \frac1t \expect[y]{\norm{x - \expect[x]{x\st y}}^2\st x}.
\end{equation}
Effective dimensionality can be equivalently defined directly from the energy. Consider the forward diffusion which progressively adds more noise to an image $x$, blurring the probability landscape. At time $t$, we can define an effective energy $\expect[y]{U(y,t) \st x}$. Its rate of change with $t$ captures the effective dimensionality around $x$. Intuitively, if this landscape is locally a $\dim$-dimensional subspace, then adding more noise causes probability to diffuse in the $d-\dim$ normal ``off-manifold'' directions, spreading over a volume $\sim t^{(d - \dim)/2}$. Taking the logarithm, we obtain $\expect[y]{U(y,t) \st x} \sim (d-\dim)\frac12\log t$. Thus, we can equivalently define
\begin{equation}
    \label{eq:dim_energy}
    \dim(x,t) = d - \partial_{\frac12\log t} \expect[y]{U(y,t) \st x}.
\end{equation}
We show in \Cref{app:dimensionality} that the definitions in \cref{eq:dim_mse,eq:dim_energy} are equivalent, unifying two seemingly distinct points of view underlying these dimensionality measures \citep{wu-verdu2011mmse-dimension,MohanKadkhodaie19b,tempczyk2022lidl-dimensionality,stanczuk2022dimension-diffusion,kamkari2024geometric-dimension,horvat2024gauge-freedom-conservativity}.



\paragraph{Numerical results.}
We show in \Cref{fig:dimensionality} the behavior of the effective energy and dimensionality as a function of the noise level. As the noise level increases, the log probability initially remains constant, indicating that probability is uniformly spread in a ball around $x$. The probability eventually decreases as the diffusion radius becomes larger than the local support size, and all lines converge to the mean (the negative entropy), consistent with the asymptotic Gaussian behavior of the energy. The effective dimensionality (\cref{fig:dimensionality}, right) vanishes at large noise levels (images have a compact support, which behaves like a point at sufficiently large noise levels) but increases as the noise level is reduced, eventually reaching the ambient dimensionality $d$ (our model has a continuous non-zero density everywhere by construction). In-between these two extremes, there exists a sizable range of scales $t \in [{10}^{-9}, {10}^{-2}]$ where almost the entire range of dimensionalities coexist. In particular, higher-probability images have lower-dimensional neighborhoods, even at relatively small scales, while lower-probability images have nearly full-dimensional neighborhoods, even at relatively large scales. This result empirically confirms the point raised in \Cref{subsec:distribution}: the local probability mass around high and low probability images are of the same order, despite the enormous gap between their probability values. The local high dimensionality of the low-probability images corresponds to a significantly larger volume occupied by their local neighborhood. 
We note two potential limits to these estimates. First, the quantization of pixel values into bins of size $1/256$ limits the resolution to $t \sim {10}^{-5}$, possibly explaining the momentary decrease in dimensionality as $t$ decreases. Second, our energy model was only trained on noise levels down to $t_{\min} = {10}^{-9}$.  Thus, energies at smaller values of $t$ are solely determined by model inductive biases, which favor a constant energy.


\section{Discussion}

We have developed a novel framework for estimating the log probability (energy) of a distribution of images from observed samples. The estimation method is simple and robust, and converges to a stable solution with relatively small amounts of data (in our examples, $100,000$ images of size $64 \times 64$ suffices). The framework leverages the tremendous power of generative diffusion models, relying on networks trained to estimate the score of the data distribution by minimizing a denoising objective. We augment this with a secondary objective that ensures consistency of the model across noise levels, and an architecture constructed from an existing denoiser so as to preserve its inductive biases. We validate our model by verifying that it achieves denoising performance equal to or surpassing the original denoiser, and NLL values comparable to the state of the art. We note that it is straightforward to extend our approach to conditional models: given conditioning information, the network needs only to compute the conditional energy function. The expression of losses $\ell_{\rm DSM}$ and $\ell_{\rm TSM}$ would be unchanged, as well as the energy architecture.

We mention two important limitations of our work. First, our energy model has a roughly quadrupled training time compared to the base score network, due to the use of double back-propagation. We believe this can be improved through architecture improvements, specialized auto-differentiation functions, or through the use of sliced score matching \citep{song2020sliced}. But in any case, the additional training time is offset by the dramatic efficiency gains in computing energy values. Another limitation of our approach is that we cannot provide any theoretical guarantees that minimizing our objective leads to a good approximation of the true energy. We expand on this question in \Cref{app:no_kl_control}.

Our learned energy model provides new insights into the geometrical properties of the space of photographic images. Notably, we observe a lack of concentration of measure, with log probability values varying over a wide range and following a Gumbel distribution. As a limit distribution for the maximum, rather than the sum, of i.i.d.\ random variables, we expect that this might arise in a broad class of high-dimensional distributions. We also introduced a novel image- and scale-dependent measure of effective dimensionality which unifies two separate points of view developed in the literature. We demonstrate that different image neighborhoods display both high and low dimensionality over a wide range of scales. These results challenge simple interpretations of the manifold hypothesis. Moreover, the estimated dimensionalities remain too high to explain how the curse of dimensionality is avoided. This suggests that the energy landscape of natural images has additional undocumented geometrical regularity.




\section*{Acknowledgments}

FG thanks Louis Thiry for starting this research direction \citep{thiry2024classification}. We also thank Joan Bruna and Pierre-Étienne Fiquet for inspiring discussions. We gratefully acknowledge the support and computing resources of the Flatiron Institute (a research
division of the Simons Foundation).


\bibliographystyle{plainnat}
\bibliography{refs}

\appendix

\clearpage
\section{Discussions on density estimation}

\subsection{On comparison of NLL values}
\label{app:nll}

We identify three issues that arise in estimation of NLL values, each of which limits the precision with which they can be compared.

A first issue concerns the selection and pre-processing of the training data. There are two different versions of ImageNet at $64 \times 64$ resolution: one introduced in \citet{van-den-oord2016pixel-cnn}, no longer available, and a second which uses anti-aliasing during downsampling (reducing entropy and thus NLL), introduced in \citet{chrabaszcz2017downsampled-imagenet}. This change in the dataset accounts for variations of about 0.3 bits/dimension \citep{zheng2023improved-ml-ode}. Similarly, the data augmentations used can both increase entropy (e.g., random flips) or reduce it (e.g., center crops, or resizing operations with anti-aliasing). 

A second point is that NLL estimates must be computed on a discrete probability model, and NLL estimates for continuous models are dependent on the method used to discretize the distribution. The simplest option is to assume that the probability is uniform within quantization bins. Discrete probabilities are then computed as a product of the continuous probability with the volume of the quantization bin, which corresponds to an additive shift of 8 bits in the NLL for image data. This can be enforced by adding uniform noise to the data, a technique known as ``uniform dequantization'' \citep{ho2019flow++}, which leads to an upper bound on the NLL of the corresponding discrete model \citep{theis2016evaluation-generative-models}. These differences can account for variations of 0.1 to 0.5 bits/dimension \citep{kong-ver-steeg-info-th-diffusion}. A more sophisticated variational dequantization procedure can improve results by an additional 0.1 bits/dimension \citep{ho2019flow++,song2021maximum}. In contrast, directly modeling the discrete data with appropriate methods can lead to reductions in NLL of more than 2 bits/dimension \citep{bhattacharya-weissman2025itdpdm-poisson}.

Finally, a third point is that depending on the method, the NLL should be interpreted differently. Classically, $p_\theta$ is a normalized probability distribution, typically obtained through a change-of-variable formula (as in flow-based models), so that up to an unknown additive constant (the entropy of the data), the NLL directly evaluates the KL divergence $\kl{p}{p_\theta}$. For some other models (such as VAEs), the NLL is not directly tied to a probabilistic model and is instead evaluated through a variational lower bound, leading to upper bounds on $-\log p(x)$ for each $x$. The NLL however remains an upper bound on the entropy of the data. The MSE-based formulas of \citet{kong-ver-steeg-info-th-diffusion} also fall in this category when using a (necessarily suboptimal) network denoiser. Lastly, approximately normalized energy-based models such as ours or CDM \citep{yadin2024classification} compute estimates of $-\log p(x)$ that are neither lower nor upper bounds, so that NLL can only be interpreted as an estimate of the entropy of the data.

As a result of these differences, NLL values of different methods are often not directly comparable. Here, we aim to learn an accurate \emph{continuous} probability model of image distributions rather than obtaining the best upper bound on the \emph{discrete} entropy of the data.

When evaluating previously published NLL results, we found that these details were often not provided, and we thus had to make assumptions in compiling \Cref{tab:nll}. The ``anti-aliasing'' column refers to the version of ImageNet64 used, we assumed that articles that cite \citet{van-den-oord2016pixel-cnn} for the dataset made use of the aliased version. We assume no data augmentation if none is mentioned (for TarFlow \citep{zhai2024tarflow}, the authors mention performing center crops of the data, but an inspection of their code indicates that it has been disabled for ImageNet64). The ``discreteness'' column refers to the nature of the probability model and the potential conversion from continuous to discrete probabilities (``discrete'' for discrete probability models, ``continuous'' for an additive shift of $8$ bits, and ``uniform'' for uniform dequantization by adding uniform noise). We assume a continuous model if no dequantization is mentioned (FM \citep{lipman2023flow-matching} and NFDM \citep{bartosh2024neural-nfdm} use a slightly different notion of uniform dequantization which improves NLL by 0.04 bits/dimension). The ``type'' column refers to the nature of the LLM estimate (``normalized'' for an exactly normalized probability model coming from a flow-based model, ``upper bound'' for variational lower bounds on log probability, and ``estimate'' for other approaches). Finally, we indicate those methods that use a single neural function evaluation (NFE) to compute log probability in the ``single NFE'' column.

\subsection{Background on density computation in diffusion models}
\label{app:density_evaluation_dm}

We review several methods of estimating log probabilities from diffusion models that have appeared in the literature.

In early publications on diffusion methods, probabilities are computed with the so-called probability flow ODE \citep{song2020score}. This corresponds to the distribution of samples generated by the backward ODE. Given a large noise level $t_{\max}$, the backward ODE solves the equation
\begin{equation}
    \label{eq:ode}
    -\frac{\diff x_t}{\diff t} = -\frac12 \nabla_y U_\theta(x_t,t),
\end{equation}
backwards in time from $x_{t_{\max}} \sim \normal(0, t_{\max}\Id)$ and produces an approximate sample $x = x_0 \sim p_{\rm ODE}$. The log probability of this sample can be calculated as \citep{song2020score}
\begin{equation}
    \label{eq:energy_ode}
    -\log p_{\rm ODE}(x) = \frac{\norm{x_{t_{\max}}}^2}{2t_{\max}} + \frac d2 \log\paren{2\pi t_{\max}} - \frac12 \int_0^{t_{\max}} \Delta_y U_\theta(x_t,t) \diff t.
\end{equation}
Note that \cref{eq:energy_ode} is also valid for arbitrary test points $x$, in which case the ODE (\ref{eq:ode}) needs to be solved \emph{forward} in time from $x_0=x$ at $t=0$ to $t=t_{\max}$. \Cref{eq:energy_ode} requires estimating the divergence of the score (the Laplacian of the energy), which is typically approximated with the Hutchinson trace estimator \citep{hutchinson1989stochastic}.

Another approach is to use a variational bound as in \citet{song2021maximum,kingma2021variational-vdm,kong-ver-steeg-info-th-diffusion}. This variational bound arises from the exact identity \citep{kong-ver-steeg-info-th-diffusion}
\begin{equation}
    \label{eq:energy_mmse}
    -\log p(x) = \expect[y]{U(y,t_{\max}) \st x} - \int_0^{t_{\max}} \paren{td - \expect[y]{\norm{x - \expect[x]{x \st y}}^2 \st x}} \frac{\diff t}{2t^2}.
\end{equation}
The first term is the effective energy at $t=t_{\max}$, which is equivalent to $\frac d2\log\paren{2\pi\expe t_{\max}}$ as $t_{\max} \to \infty$. The second term features the optimal mean squared error over noise levels $t \in [0, t_{\max}]$. Note that the integrand can be rewritten as $(d - \dim(x,t)) \diff\paren{\frac12 \log t}$, and \cref{eq:energy_mmse} can thus be derived by integrating the two equivalent definitions of effective dimensionality (\ref{eq:dim_mse}) and (\ref{eq:dim_energy}) (see \Cref{app:dimensionality}).
\Cref{eq:energy_mmse} naturally leads to an upper-bound on the negative log probability of the data when replacing the optimal denoiser $\expect[x]{x \st y,t}$ with the denoiser derived from the Miyasawa-Tweedie expression, $y - t\nabla_y U_\theta(y,t)$:
\begin{equation}
    \label{eq:energy_mse}
    -\log p_{\rm MSE}(x) = \frac d2\log\paren{2\pi\expe t_{\max}} - \int_0^{t_{\max}} \paren{td - \expect[y]{\norm{x - y + t\nabla_y U_\theta(y,t)}^2 \st x}} \frac{\diff t}{2t^2}.
\end{equation}
This framework can be generalized to other noise distributions \citep{guo-verdu2013information-estimation} such as Poisson noise, which is more adapted to discrete distributions \citep{bhattacharya-weissman2025itdpdm-poisson}.

Finally, it was recently observed in \citet{skreta2024ito-density-estimation,karczewski2024diffusion-cartoonists} that a cheaper unbiased stochastic estimator of this bound can be obtained with the Itô formula. Given a realization $(x_t)_{t \in \RR_+}$ of the forward SDE (Brownian motion) $\diff x_t = \diff w_t$ started at $x_0 = x \sim p(x)$, then
\begin{equation}
    -\log p(x) = U(x_{t_{\max}}, t_{\max}) - \int_0^{t_{\max}} \paren{\inner{\nabla_y U(x_t,t), \diff x_t} - \frac12\norm{\nabla_y U(x_t,t)}^2\diff t}.
\end{equation}
When $t_{\max} \to \infty$, one can again use $U(x_{t_{\max}},t_{\max}) \sim \frac{\norm{x_{t_{\max}}}^2}{2t_{\max}} + \frac d2 \log\paren{2\pi t_{\max}}$. Replacing the unknown true energy $U$ in the integrand with a model $U_\theta$ leads to a biased stochastic estimate of the log probability:
\begin{equation}
    \label{eq:energy_sde}
    -\log p_{\rm SDE}(x) = \frac{\norm{x_{t_{\max}}}^2}{2t_{\max}} + \frac d2\log\paren{2\pi t_{\max}} - \int_0^{t_{\max}} \paren{\inner{\nabla_y U_\theta(x_t,t), \diff x_t} - \frac12\norm{\nabla_y U_\theta(x_t,t)}^2\diff t}.
\end{equation}
\citet{karczewski2024diffusion-cartoonists} show that averaging over SDE trajectories started at the same $x_0 = x$ leads to an upper bound on the NLL, in fact recovering the one in \cref{eq:energy_mse}.

In summary, the ODE gives deterministic probabilities exactly corresponding to the corresponding generative model (\cref{eq:energy_ode}), while for the SDE one can choose between a denoising-based upper bound (\cref{eq:energy_mse}) or a cheap stochastic estimator of it (\cref{eq:energy_sde}).
These evidently similar approaches are equivalent if and only if the model is consistent across noise levels (i.e., satisfies the diffusion equation). In practice, the choice of estimation method can lead to variations of up to 0.5 bits/dimension \citep{kong-ver-steeg-info-th-diffusion}, see in particular \citet{karczewski2024diffusion-cartoonists} for a careful study). 
These approaches also apply to our model, in addition to the direct evaluation of $U_\theta(x,t=0)$. While this latter estimate does not correspond to a generative model (like the ODE) or a variational bound (like the SDE), its main advantage is its computational efficiency, as it does not require integrating over noise levels and gives deterministic values without averaging over noise realizations (for a divergence term or mean squared error).

\subsection{On the theoretical justification of dual score matching}
\label{app:no_kl_control}
A limitation of our approach is that we offer no theoretical guarantees that minimizing our objective leads to a good approximation of the energy.
It would be desirable to show that our training objective (assuming infinite training data) quantitatively controls the distance between the learned energy and the data energy. 
This could be achieved by showing that the joint distribution $p(y,t)$ (with $\log t$ uniformly distributed in $[\log t_{\min}, \log t_{\max}]$) satisfies a Poincaré inequality, i.e.,
\begin{equation}
    \label{eq:poincare}
    \vvar{U_\theta(y,t) - U(y,t)} \leq C \expect{\norm{\nabla_y U_\theta(y,t) - \nabla_y U(y,t)}^2 + \parenn{\partial_t U_\theta(y,t) - \partial_t U(y,t)}^2},
\end{equation}
for some constant $C$ that is not too large.
While we conjecture that such a result holds (or a variant, such as when considering the distribution of $(y/\sqrt t, \log t)$, to match our noise level weighting), note that this is weaker than a control in the Kullback-Leibler divergence. Replacing $\vvar{U_\theta(y,t) - U(y,t)}$ with $\kl{p}{p_\theta}$ in \cref{eq:poincare} would require that the \emph{model} distribution $p_\theta(y,t)$ satisfies a log-Sobolev inequality, which could only be enforced with specific architectures. As a result, there is no control of the learned energy outside the support of the data distribution, so that out-of-distribution detection may be unreliable. It also implies our model may not be exactly normalized. Our NLL calculations are thus only approximate (although computationally cheap). These limitations are common to all current score-based and unnormalized energy-based approaches, including the related work of \citet{yadin2024classification}.

\subsection{On frequency estimation with density models}

Here, we describe a counterintuitive property of density models in high dimensions which poses a challenge for estimating frequencies. We also refer the interested reader to \citet{theis2016evaluation-generative-models}, which offers related observations.

Consider a mixture of two uniform distributions on two compact sets (classes) $C_1$ and $C_2$ with respective frequencies $f_1$ and $f_2$. The probability density is then constant on each class, with value $p_i = \frac{f_i}{V_i}$ where $V_i$ is the volume of $C_i$. The volume $V_i$ typically scales exponentially with $d$, $V_i \sim r_i^d$ where $r_i$ is the characteristic size of $C_i$. The energy values on each class are then 
\begin{equation}
    U_i = -\log p_i = d \log r_i - \log f_i.
\end{equation} 
In high dimensions $d \gg 1$, the energy values are dominated by the volume and only weakly depend on the frequencies of each class. 

A concrete example of this phenomenon can be seen in the mixture of two Gaussian distributions considered in \Cref{fig:single_vs_dual_sm}, where the two classes correspond to the two spheres of radius $\sqrt d \sigma_i$ for $i\in \{1,2\}$. The typical values of the energy $U_i = \frac d2 \log\paren{2\pi\expe \sigma_i^2} - \log f_i$ are dominated by the entropy of the corresponding Gaussian. 

The implications of this observation for high-dimensional density models are twofold. First, estimated densities are dominated by volumes of typical sets (entropies of the mixture components), more than frequencies of different categories (one-dimensional marginals). (Note that the latter is easily learned from data, while estimating the former is a much more challenging task.) This observation explains empirical reports that energy models may assign higher probability to out-of-distribution samples \citep{nalisnick2019deep,karczewski2024diffusion-cartoonists}.
Second, estimating energy up to a small relative error is not sufficient to capture observed frequencies. For our ImageNet64 model ($d = 4096$), a relative error of $\sim 0.1$ dB/dimension (as estimated from the generalization experiment in \Cref{fig:generalization}) is negligible compared to typical energy differences of $\sim 10$ dB/dimension, but the required precision to estimate frequencies with a relative accuracy of $10\%$ is $1/d = 2\times {10}^{-4}$ dB/dimension.

\section{Proofs and derivations}

\subsection{Time score matching (proof of \cref{eq:time_score_identity})}
\label{app:time_score_matching}

We start from the expression of the energy (negative log probability) of noisy data:
\begin{align}
    U(y,t) &= -\log p(y|t) , \\
    p(y|t) &= \int p(x) p(y|x,t) \diff x . 
    \label{eq:awgn}
\end{align}
Differentiating the energy with respect to $t$ yields
\begin{align}
    \partial_t U(y,t) &= -\frac{\partial_t p(y|t)}{p(y|t)} \\
    &= -\frac{\int p(x) p(y|x,t) \partial_t \log p(y|x,t) \diff x}{p(y|t)} \\
    &= \expect[x]{-\partial_t \log p(y|x,t) \st y,t}.
    \label{eq:time_score_cond_exp}
\end{align}
Note that the derivation exactly matches that of the Miyasawa-Tweedie identity by replacing $\partial_t$ with $\nabla_y$, and does not make any assumptions about the form of $p(y|x,t)$.  Restricting to additive Gaussian noise, $-\log p(y|x,t) = \frac1{2t}\norm{y-x}^2 +\frac d2 \log(2\pi t)$, and substituting into \cref{eq:time_score_cond_exp} gives the ``time score-matching'' identity of \cref{eq:time_score_identity}:
\begin{equation}
    \partial_t U(y,t) = \expect[x]{\frac{d}{2t} - \frac{\norm{y-x}^2}{2t^2} \st y}.
\end{equation}

\subsection{Energy architecture}
\label{app:energy_architecture}

Suppose that the energy network is defined in terms of a score network $s_\theta(y,t)$ as $U_\theta(y,t) = \frac12 \inner{y, s_\theta(y,t)}$, where the score network is assumed conservative and homogeneous. Conservativity means that there exists a scalar function $\phi$ such that $s_\theta(y,t) = \nabla_y \phi(y,t)$, which implies that the Jacobian $\nabla_y s_\theta(y,t) = \nabla^2_y \phi(y,t)$ is symmetric. The homogeneity property requires that for all $\lambda \geq 0$, $s_\theta(\lambda y, t) = \lambda s_\theta(y,t)$. Differentiating with respect to $\lambda$ and setting $\lambda = 1$ yields
\begin{equation}
    \nabla_y s_\theta(y,t) y = s_\theta(y,t).
\end{equation}

We now calculate the gradient of the energy network:
\begin{align}
    \nabla_y U_\theta(y,t) = \frac12 \paren{s_\theta(y,t) + \nabla_y s_\theta(y,t)\trans y} = s_\theta(y,t),
\end{align}
using the conservative and homogeneity properties to derive that $\nabla_y s_\theta(y,t) y = \nabla_y s_\theta(y,t)\trans y = s_\theta(y,t)$. Note that even if $s_\theta$ is not conservative, then it still holds that $\nabla_y U_\theta(y,t) = \frac12\paren{\nabla_y s_\theta(y,t) + \nabla_y s_\theta(y,t)\trans} y$, which can be interpreted as a symmetrization of the Jacobian of $s_\theta$.

We also remark that if $s_\theta$ is homogeneous, then $U_\theta$ is quadratically homogeneous ($U_\theta(\lambda y, t) = \lambda^2 U_\theta(y, t)$). Note that this does not correspond to enforcing (asymmetric) Gaussian one-dimensional marginals $\langle y, u\rangle$. Rather, this enforces that the distribution of $\langle y, u\rangle$ conditioned on the orthogonal projection $y - \langle y, u\rangle u/\norm{u}^2$ is (asymmetric) Gaussian. 

\subsection{Effective dimensionality (equivalence between \cref{eq:dim_mse,eq:dim_energy})}
\label{app:dimensionality}

We start by calculating the time derivative of the effective energy $\expect[y]{U(y,t) \st x}$. We have
\begin{align}
    \partial_t \expect[y]{U(y,t) \st x}
    &= \partial_t \int p(y|x,t) U(y,t) \diff y \\
    &= \int \paren{\partial_t p(y|x,t) U(y,t) + p(y|x,t) \partial_t U(y,t)} \diff y.
    \label{eq:eff_dim_two_terms}
\end{align}
The first term is the derivative with respect to variance of a Gaussian distribution $\normal(x, t\Id)$. It satisfies the diffusion equation: 
\begin{equation}
    \label{eq:eff_dim_term1}
    \partial_t p(y|x,t) = \frac12 \Delta_y p(y|x,t).
\end{equation}
Similarly, for the second term, we use the fact that $U(y,t) = -\log p(y|t)$ where $p(y|t)$ also satisfies the diffusion equation (as can be seen from the Fokker-Planck equation in the variance exploding case \citep{song2020score}):
\begin{align}
    \partial_t U(y,t) &= -\frac{\partial_t p(y|t)}{p(y|t)} \\
    &= -\frac{\Delta_y p(y|t)}{2p(y|t)} \\
    &= \frac1{2p(y|t)} \nabla_y\cdot \paren{p(y|t) \nabla_y U(y|t)} \\
    &= \frac12 \Delta_y U(y|t) - \frac12 \norm{\nabla_y U(y|t)}^2.
    \label{eq:eff_dim:term2}
\end{align}
This equation appeared in \citet{lai-ermon2023score-fokker-planck,bruna-han2024provable-posterior-sampling} and is a special case of a Hamilton-Jacobi equation \citep{evans2022pde}. 

Combining \cref{eq:eff_dim_term1,eq:eff_dim:term2} into \cref{eq:eff_dim_two_terms} and integrating by parts twice, we have
\begin{align}
    \partial_t \expect[y]{U(y,t) \st x}
    &= \int p(y|x,t) \paren{\Delta_y U(y,t) - \frac12 \norm{\nabla_y U(y|t)}^2} \diff y.
    \label{eq:eff_dim_eff_energy_rate}
\end{align}
We recognize the expression of the mean squared error as given by combining Miyasawa-Tweedie with Stein's unbiased risk estimate (SURE). Indeed,
\begin{align}
    \expect[y]{\norm{x - \expect[x]{x \st y}}^2 \st x} 
    &= \expect[y]{\norm{x - y + t\nabla_y U(y,t)}^2 \st x} \\
    &= \expect[y]{\norm{x - y}^2 + 2t\inner{x - y, \nabla_y U(y,t)} + t^2 \norm{\nabla_y U(y,t)}^2 \st x} \\
    &= td + t^2\expect{-2\Delta_y U(y,t) + \norm{\nabla_y U(y,t)}^2 \st x},
    \label{eq:eff_dim_sure}
\end{align}
where we have used Stein's lemma in the last step. 
We finally combine \cref{eq:eff_dim_eff_energy_rate,eq:eff_dim_sure} to obtain
\begin{equation}
    \label{eq:eff_dim_definition}
    \frac1t \expect[y]{\norm{x - \expect[x]{x \st y}}^2 \st x} = d - 2t \partial_t \expect[y]{U(y,t) \st x}.
\end{equation}
It is convenient to rewrite $2t \partial_t = \partial_{\frac12 \log t}$ as $\diff \paren{\frac12 \log t} = \frac{\diff t}{2t}$.

We define the common value in \cref{eq:eff_dim_definition} to be $\dim(x,t)$. It is related (but not equal) to dimensionality measures estimated from the singular values of the Jacobian of a denoiser \citep{MohanKadkhodaie19b,horvat2024gauge-freedom-conservativity}. The limit of $\dim(x,t)$ when $t \to 0$ has appeared in the literature under the name of local intrinsic dimensionality using (approximate) likelihood (LIDL) \citep{tempczyk2022lidl-dimensionality,stanczuk2022dimension-diffusion,kamkari2024geometric-dimension}. Its average over images $x$ is equal to the MMSE dimension of \citet{wu-verdu2011mmse-dimension}.

\section{Experimental details}
\label{app:all_exp_details}

\subsection{Energy network architecture}
\label{app:architecture_details}

\paragraph{UNet architecture.}
Our UNet architecture $s_\theta$ is composed of $3$ encoder blocks, a middle block, and $3$ decoder blocks. Each block is itself composed of $3$ layers, each a sequence of bias-free convolution, normalization, and non-linearity, for a total of $21$ layers. The first convolutional layer of each encoder block but the first and the middle block has a stride of $2$ (downsampling, in both vertical and horizontal directions), and the last convolutional layer of each decoder block (except for the last one and the middle block) is transposed with a stride of $2$ (upsampling). The output of each encoder block is concatenated to the input of the corresponding decoder block (which comes from the output of the corresponding encoder block, via a ``skip connection''). The number of channels is doubled in each block, starting from a base value of $64$ at the coarsest scale. We replace ReLUs with GeLUs to ensure that $\nabla_y U_\theta$ is differentiable.  Thus, $U_\theta$ is only approximately quadratically homogeneous. As a result, and because training does not result in an exactly conservative $s_\theta$, $\nabla_y U_\theta(y,t)$ should be computed directly as opposed to using $s_\theta$. 

\paragraph{Normalization.}
We also replace batch normalizations, whose behavior during the backward pass is incompatible with a second back-propagation, with a homogeneous version of instance normalization \citep{ulyanov2016instance-normalization}. If the input $x$ consists of $C$ channels $(x_c)_{1\leq c \leq C}$, and its spatial mean is $\mu(x) \in \RR^C$, each channel is normalized according to
\begin{equation}
    x_c \mapsto \sqrt{\frac{\|x - \mu(x)\|^2 + \varepsilon}{ C \|x_c - \mu(x_c)\|^2 + \varepsilon}} (x_c - \mu(x_c)).
\end{equation}
Up to a small $\varepsilon > 0$ parameter for numerical stability, this ensures that after normalization all channels $x_c$ have equal norms while preserving the global norm $\|x\|$. This normalization layer is also homogeneous when $\varepsilon = 0$, as the spatial mean is estimated from the input $x$. This normalization is followed by a learned rescaling of each channel, $x_c \mapsto \gamma_c x_c$, where $\gamma \in \RR^C$ is learnable.

\paragraph{Noise level conditioning.}
The noise variance $t$ is also an input of $s_\theta$. As is standard in diffusion models \citep{nichol-dhariwal2021improved-ddpm}, a time embedding $e(t) \in \RR^{256}$ is computed with Fourier features $\cos(\omega_k t)$, $\sin(\omega_k t)$ (we use $32$ frequencies $(\omega_k)_k$ that are linearly spaced in the log domain and ranging from $1/t_{\max}$ to $1/t_{\min}$) followed by a shallow MLP. This time embedding $e(t)$ is then used to condition the output of each normalization layer via gain control: $x_c \mapsto ((1 + \inner{w_c, e(t)}) x_c)$ where $w_c$ is a learned layer- and channel-dependent vector $\in \RR^{256}$.

\subsection{Training hyper-parameters}
\label{app:training_details}

To summarize \Cref{sec:objective}, the dual score matching training objective is
\begin{equation}
    \ell(\theta) = \expect[x,z,t]{\norm{\sqrt{\frac td} \nabla_y \energy(y,t) - \frac{z}{\sqrt d}}^2 + \paren{\frac td \partial_t \energy(y,t) - \frac12\paren{1 - \frac{\norm{z}^2}{d}}}^2},
\end{equation}
where $x\sim p(x)$ is the data distribution, $z \sim \normal(0, \Id)$ is the noise, $y = x +\sqrt t z$ is the noisy measurement, and $\log t \sim \mathcal U(\log t_{\min}, \log t_{\max})$. In our experiments we use $t_{\min} = {10}^{-9}$ and $t_{\max} = {10}^3$, and the training image intensities are rescaled to have values in $[0,1]$.

We use the ImageNet64 dataset \citep{chrabaszcz2017downsampled-imagenet}, with (only) horizontal flips as data augmentation. The models used in \Cref{tab:denoising_performance,tab:nll} and for the generalization experiment in \Cref{fig:generalization} are trained on color images, while the model used in \Cref{fig:energy_distribution,fig:energy_vs_sparsity,fig:dimensionality} is trained on grayscale images. Pixel values are rescaled to $[0,1]$ by dividing by $255$. All models are trained for $1\text{M}$ steps, with a batch size of $128$. We use the Adam optimizer with default parameters and an initial learning rate of $0.0005$ (except for the generalization experiments which used a learning rate of $0.0002$) that is halved every $100,000$ steps. All models are trained on a single NVIDIA H100 GPU, which takes about $5$ days for ImageNet64.

\subsection{Additional details}
\label{app:addtional_details}

\paragraph{Gaussian scale mixture example (\Cref{fig:single_vs_dual_sm}).}
We generate $n=100,000$ samples from a mixture of two Gaussian distributions, $\frac12\normal(0, \sigma_1^2\Id) + \frac12 \normal(0, \sigma_2^2\Id)$, with $\sigma_1=1$ and $\sigma_2=4$, in dimension $d = 1,000$. The true (normalized) energy is given by
\begin{equation}
    U(y,t) = -\log\paren{\summ i2 \expe^{-\frac{\norm{y}^2}{2\paren{\sigma_i^2 + t}} - \frac d2 \log\paren{2\pi\paren{\sigma_i^2 + t}} - \log 2}}.
\end{equation}
We parameterize the energy as a mixture of quadratics:
\begin{equation}
    U_\theta(y,t) = -\log\paren{\summ i2 \expe^{- a_i(t) \norm{y}^2 - b_i(t)}},
\end{equation}
where the functions $a_i, b_i$ are computed by a $5$-layer MLP with a hidden dimension of $256$ that takes $\log(t + t_{\min})$ as input. This network is trained either with single (space) score matching or with dual score matching, both across noise levels $t \in [t_{\min}, t_{\max}]$. Training is otherwise similar to the ImageNet64 experiments (see \Cref{app:training_details}), for $20,000$ training steps with a batch size of $512$ and an initial learning rate of $0.0001$, over noise levels from $t_{\min} = {10}^{-2}$ to $t_{\max} = {10}^2$. We note that the energy learned after single score matching training is stochastic: as the relative energy between the two mixture components is not constrained by the data, its value is determined by the random initialization, so that rerunning the experiment will lead to a different value.

\paragraph{Histogram of log probabilities (\Cref{fig:energy_distribution}).}
The Gumbel fit is calculated by maximizing likelihood. The obtained parameters (in decibels/dimension) are $9.57$ for the location and $3.17$ for the scale. Equivalently, $p(x)$ follows a Fréchet distribution, with a scale parameter equal to ${9.05}^d$ and shape parameter equal to ${1.37}/d$, where $d = 64 \times 64 = 4096$ is the dimension of ImageNet64 grayscale images.

\paragraph{Computing dimensionality (\Cref{fig:dimensionality}).}
\Cref{eq:dim_mse,eq:dim_energy} provide two ways to estimate effective dimensionality from a learned energy model. If the model is exact, $U_\theta(y,t) = U(y,t)$, then they coincide (more generally, they coincide when the model satisfies the diffusion equation), but in general they only approximately coincide. For numerical stability, we found it preferable to use the version defined from the mean squared error of the underlying denoiser (\cref{eq:dim_mse}). This guarantees non-negative dimensionalities, and also has the advantage of being an upper bound on the true dimensionality (as any denoiser yields an upper bound on the minimum MSE).

\section{Additional results}
\subsection{ImageNet images according to their probability}
\label{app:more_images}

\begin{figure}[H]
    \centering
    \includegraphics[width=1\linewidth]{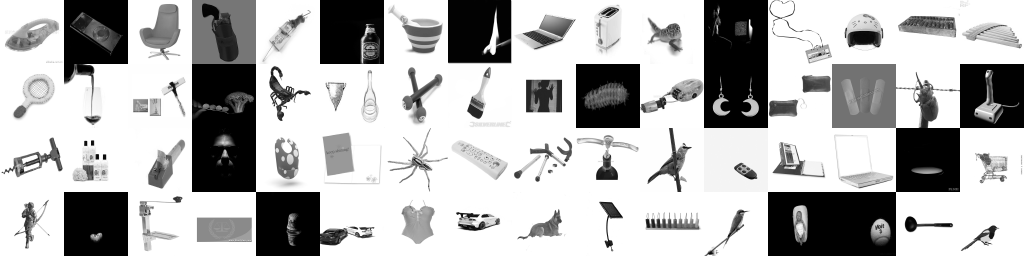}
    \caption{Highest probability images in ImageNet64 (test set).}
    \label{fig:low-energy-images}
\end{figure}

\begin{figure}[H]
    \centering
    \includegraphics[width=1\linewidth]{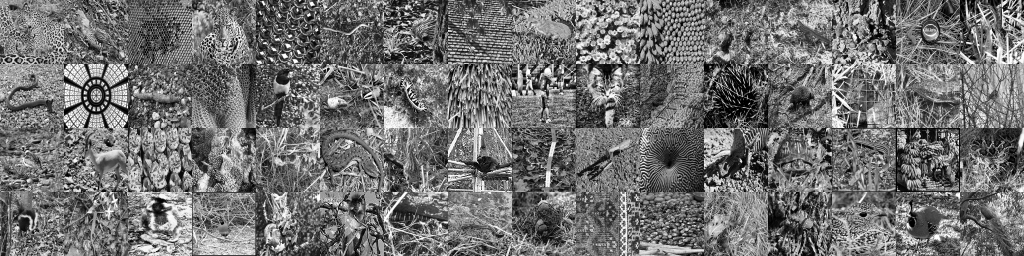}
    \caption{Lowest probability images in ImageNet64 (test set).}
    \label{fig:high-energy-images}
\end{figure}

\begin{figure}[H]
    \centering
    \includegraphics[width=1\linewidth]{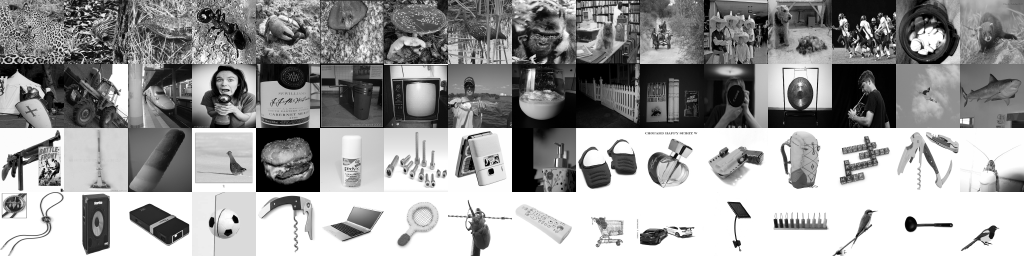}
    \caption{Example images from ImageNet64 (test set) with linearly-spaced log probabilities, ordered from low to high.}
    \label{fig:linear-energy-images}
\end{figure}

\subsection{Distribution of log probabilities on other datasets}
\label{app:more_histograms}

\begin{figure}[H]
    \centering
    \includegraphics[width=0.47\linewidth]{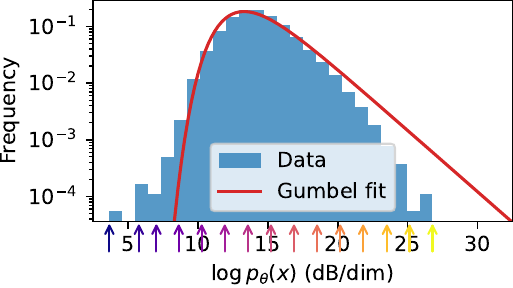}
    \hfill
    \raisebox{4pt}{\includegraphics[width=0.43\linewidth]{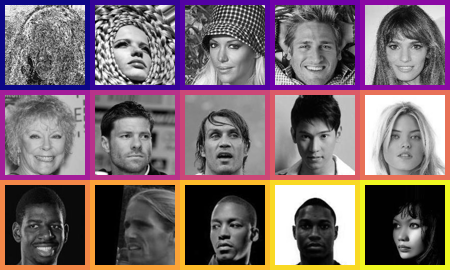}}
    \caption{
    Histogram of log probabilities of test images in the CelebA dataset (grayscale). Color-coded arrows indicate values for the example images on the right. The maximum-likelihood fit by a Gumbel distribution (red line) is perturbed by the low-probability outliers, some of which correspond to corrupted images.
}
    \label{fig:energy_distribution_celeba_grayscale}
\end{figure}

\begin{figure}[H]
    \centering
    \includegraphics[width=0.47\linewidth]{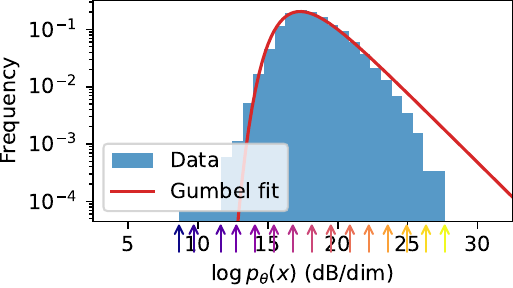}
    \hfill
    \raisebox{4pt}{\includegraphics[width=0.43\linewidth]{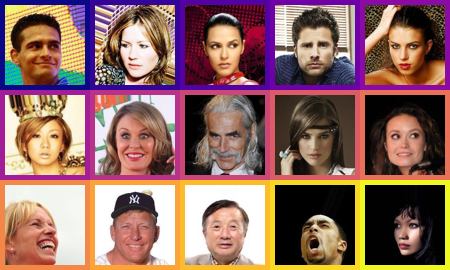}}
    \caption{
    Histogram of log probabilities of test images in the CelebA dataset (color). Color-coded arrows indicate values for the example images on the right. The maximum-likelihood fit by a Gumbel distribution (red line) is perturbed by the low-probability outliers.
}
    \label{fig:energy_distribution_celeba}
\end{figure}

\begin{figure}[H]
    \centering
    \includegraphics[width=0.47\linewidth]{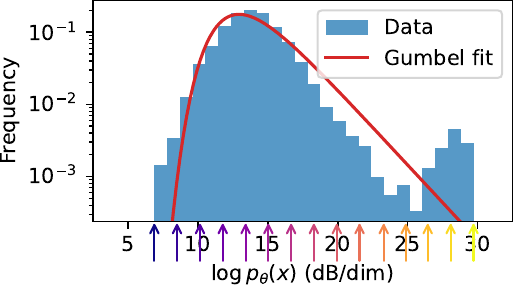}
    \hfill
    \raisebox{4pt}{\includegraphics[width=0.43\linewidth]{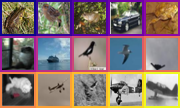}}
    \caption{
    Histogram of log probabilities of test images in the CIFAR-10 dataset. Color-coded arrows indicate values for the example images on the right. The maximum-likelihood fit by a Gumbel distribution (red line) is perturbed by high-probability outliers, which correspond to grayscale images.
}
    \label{fig:energy_distribution_cifar}
\end{figure}

To evaluate the pervasiveness of the Gumbel-distributed log probability values, we trained energy models on CelebA (grayscale and color) and CIFAR-10, with an identical architecture, objective, and hyper-parameters as described in \Cref{sec:methods} and \Cref{app:all_exp_details}. Results are presented in \Cref{fig:energy_distribution_celeba_grayscale,fig:energy_distribution_celeba,fig:energy_distribution_cifar}. In all cases, log-probability histograms share qualitative features with the Gumbel distribution, namely its skewness towards high-probability values, although the fits are less precise due to the presence of low- or high-probability outliers.

\subsection{Log probabilities of interpolated images}
\label{app:interpolation}

\begin{figure}[H]
    \centering
    \includegraphics[width=0.47\linewidth]{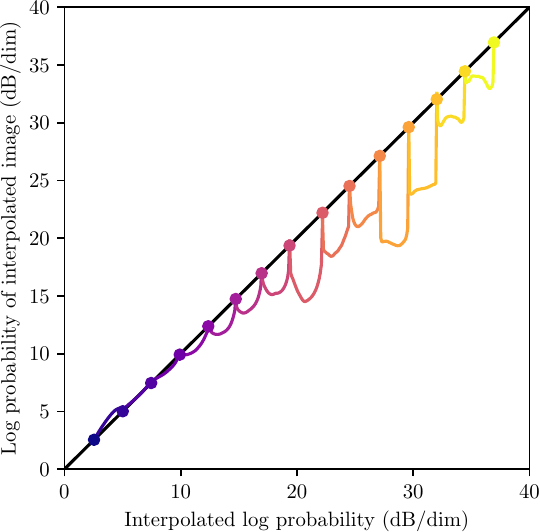}
    \hfill
    \raisebox{10pt}{\includegraphics[width=0.43\linewidth]{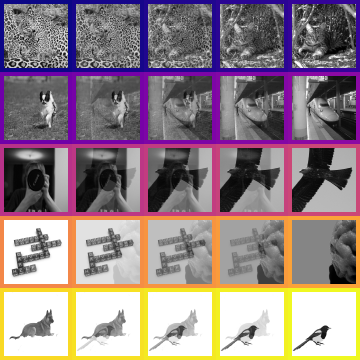}}
    \caption{
    Log probability of interpolated images. For each pair $(x_i, x_{i+1})$ of consecutive example images (see \Cref{fig:energy_distribution}), we consider a linear path between them, $\alpha\in [0,1] \mapsto (1-\alpha)x_i + \alpha x_{i+1}$ (shown in the right panel). In the left panel, we plot the log probability of the resulting images $\log p_\theta((1-\alpha)x_i + \alpha x_{i+1})$ as a function of the interpolated log probability $\alpha \log p_\theta(x_i) + (1-\alpha)\log p_\theta(x_{i+1})$. For medium- and high-probability images, this leads to a sharp decrease in probability even for $\alpha$ close to $0$ or $1$: the superposition of two natural images is not a natural image. However, for low-probability images, this surprisingly leads to a slight increase in probability: the interpolation between two densely textured images is another densely textured image but with slightly lower contrast, leading to higher probability values (as shown in \Cref{fig:energy_vs_sparsity}).
}
    \label{fig:energy_interpolation}
\end{figure}

\subsection{Optimizing log probability}
\label{app:logp_optimization}

\begin{figure}[H]
    \centering
    \includegraphics[width=\linewidth]{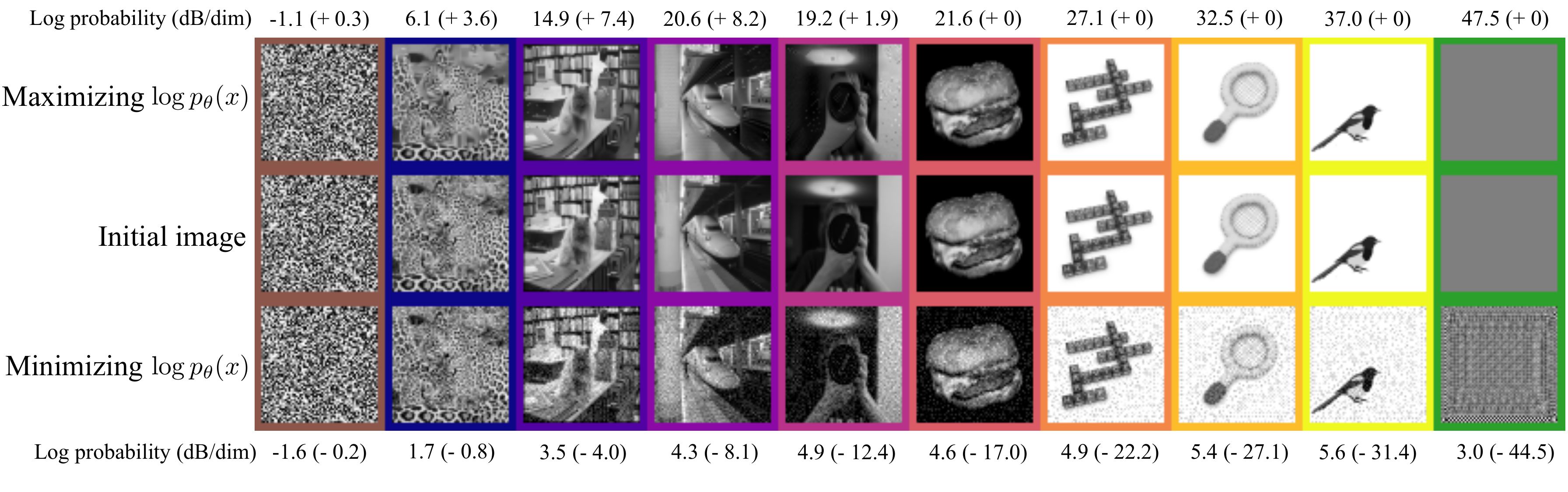}
    \vspace{-1.5em}
    \caption{
    For each initial image (middle row), we perform $1,\!000$ steps of gradient ascent (top row) or descent (middle row) on the log probability with a fixed step size of ${10}^{-2}/d$, clipping pixel intensity values to $[0,1]$ at each step. Due to numerical instabilities around high-probability images, we select the iterate with the highest or lowest probability. Resulting log probability values, and their difference with the starting value, are shown above or below the final images. Maximizing log probability leads to generally smoother images (although with some visual artifacts), except for sufficiently high-probability images, where gradient ascent fails to increase the probability. Minimizing log probability leads to noisier images, with regular geometric patterns for high-probability images.
}
    \label{fig:logp_optimization}
\end{figure}

\end{document}